\documentclass[11pt]{article}

\usepackage[final]{acl}

\usepackage{times}
\usepackage{latexsym}

\usepackage[T1]{fontenc}
\usepackage[utf8]{inputenc}

\usepackage{microtype}

\usepackage{inconsolata}

\usepackage{graphicx}
\usepackage{amsmath}
\usepackage{booktabs}
\usepackage{multirow}
\usepackage{epigraph}
\usepackage{algorithm}
\usepackage{algorithm}
\usepackage{algpseudocode}  % 替代algorithmic
\usepackage{xcolor}
\algtext*{EndFor}\algtext*{EndWhile}\algtext*{EndIf}
\usepackage{xcolor}  % for \textcolor
\usepackage{tcolorbox}
\tcbuselibrary{skins,breakable}  % 可选：增强功能

\usepackage{listings}
\usepackage{xcolor}

\usepackage{fancyvrb}

\usepackage{amssymb}  % for \checkmark
\usepackage{pifont}   % for \ding{51}, \ding{55}

\newcommand{\symbolfootnote}[2]{%
    \renewcommand{\thefootnote}{#1}% 将编号显示改为你的符号
    \footnotetext[0]{#2}% 使用 [0] 确保不增加数字计数
    \renewcommand{\thefootnote}{\arabic{footnote}}% 恢复为数字
}

\title{Knowing When to Quit: Diagnosing and \\
Training LLMs to Abort \textit{Futile Reasoning}}

\author{\textbf{Xinyan Guan\textsuperscript{1,2}},
 \textbf{Jiali Zeng\textsuperscript{3,$\dagger$}},
 \textbf{Chunlei Xin\textsuperscript{1,2}},
 \textbf{Yaojie Lu\textsuperscript{1,$\dagger$}},
 \\
 \textbf{Hongyu Lin\textsuperscript{1}},
 \textbf{Xianpei Han\textsuperscript{1}},
 \textbf{Le Sun \textsuperscript{1}},
 \textbf{Fandong Meng\textsuperscript{3}}
\\
 \textsuperscript{1}Chinese Information Processing Laboratory, Institute of Software, Chinese Academy of Sciences\\
 \textsuperscript{2}University of Chinese Academy of Sciences\\
 \textsuperscript{3}Weixin AI, Tencent Inc, China
\\
\texttt{\{guanxinyan2022,chunlei2021,luyaojie,hongyu,xianpei,sunle\}@iscas.ac.cn} \\
 \texttt{\{lemonzeng,fandongmeng\}@tencent.com} \\
}

\begin{document}
\maketitle

\symbolfootnote{$\dagger$}{Corresponding author.}

\begin{abstract}

Large language models generate computationally expensive yet semantically 
void reasoning on beyond-capability tasks, creating risks where 
plausible-sounding but incorrect derivations mislead users. We characterize 
this \textit{futile reasoning} phenomenon through systematic analysis, 
revealing universal capability overreach and systematic miscalibration 
between capability and behavior.
The dominant failure mode is 
specious reasoning, which outputs look superficially valid but contain subtle errors, escalating with task difficulty. 
To address this, we introduce 
\textbf{CaRL} (\textbf{Ca}pability-\textbf{a}ligned \textbf{R}einforcement 
\textbf{L}earning), which aligns model behavior with capability boundaries 
through reward shaping that incentivizes refusal over futile reasoning and 
hindsight refusal augmentation that converts failures into refusal supervision. 
Experiments demonstrate a substantial reduction in futile reasoning while 
preserving performance across task difficulties, effectively achieving 
capability-aligned behavior without sacrificing utility. \footnote{\url{https://github.com/icip-cas/Knowing-When-to-Quit}}

\end{abstract}

% \epigraph{\textit{``Forcing competence beyond one's capability is a critical failure.''}}{--- Mozi}

\section{Introduction}

% Large language models (LLMs), particularly characterized by their substantial number of parameters, have arisen as a promising tool for solving various tasks. Furthermore, generative reasoning model (GRM) futher inctizive reasoning capablities by test-time sacling 

% Large Language Models (LLMs) have demonstrated remarkable proficiency in complex reasoning tasks~\cite{xu2025towards,minaee2024large,zhao2023survey}, particularly when augmented with long thinking patterns. 

Large language models (LLMs), characterized by billions of parameters, have emerged as powerful tools for solving various tasks~\cite{xu2025towards,minaee2024large,zhao2023survey}. Recent advances in generative reasoning models (GRMs) further amplify their reasoning capabilities through test-time scaling~\cite{xu2025towards,huang2023towards}, unlocking the ability to solve increasingly complex reasoning problems.

However, when confronted with tasks that exceed their intrinsic capabilities, LLMs fail to recognize their boundaries.
Instead of acknowledging their ignorance, they often persist in generating output, producing outputs that superficially resemble valid solutions.
Due to the lengthy and convoluted nature of these generated traces, distinguishing valid reasoning from such hallucinations is notoriously difficult for humans~\cite{bowman2022measuring,perez2023discovering}.
Consequently, this creates a critical reliability risk: users may mistakenly trust plausible-sounding but fundamentally incorrect derivations, severely limiting the adoption of LLMs in high-reliability domains~\cite{liu2023trustworthy}.

\begin{figure}
    \centering
    \includegraphics[width=1.0\linewidth]{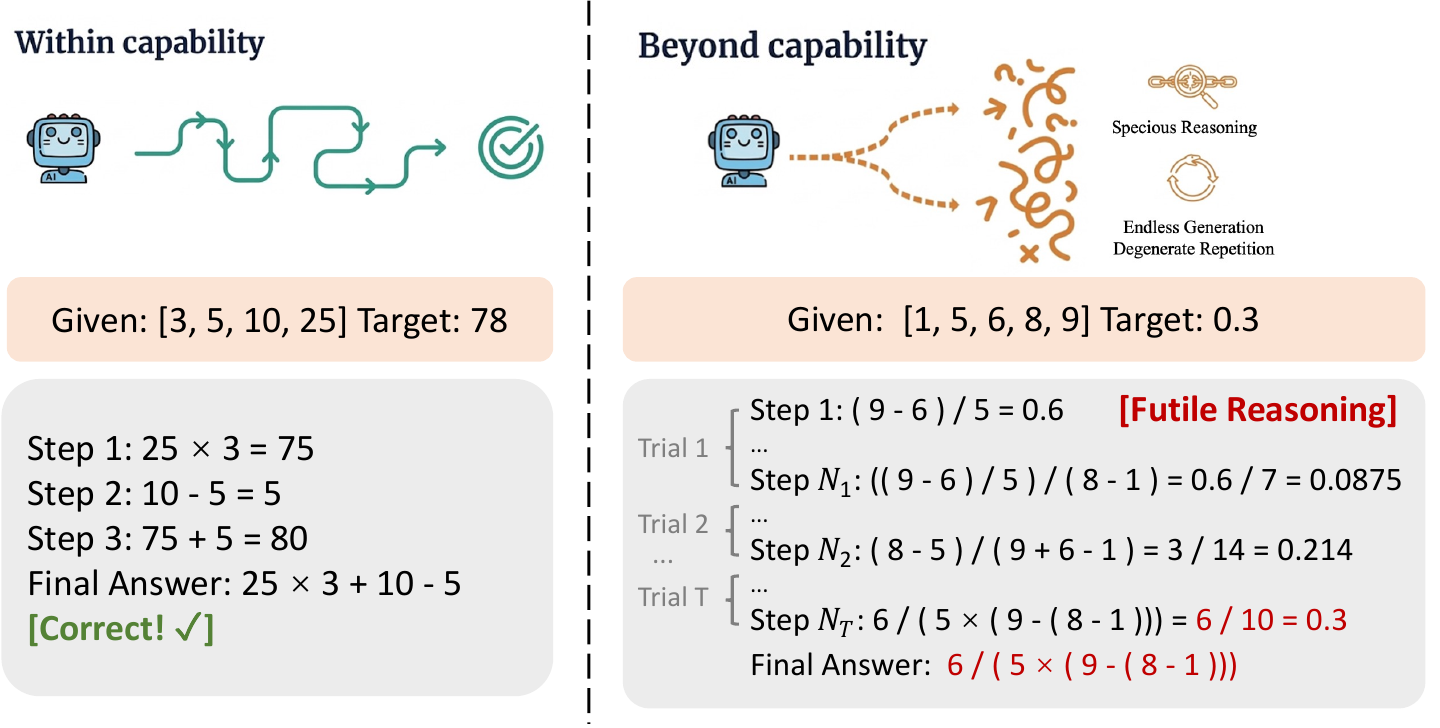}
    \caption{Illustration of \textit{\textbf{futile reasoning}} generated by DeepSeek-R1~\cite{guo2025deepseek}. When faced with problems beyond its capability, the model generates plausible-looking but fundamentally incorrect reasoning traces instead of refusing to answer.}
    \label{fig:fig1}
\end{figure}

We term this phenomenon \textbf{\textit{futile reasoning}}, referring 
to the plausibly-looking but fundamentally incorrect reasoning attempts triggered when models face problems beyond their capabilities. 
As illustrated in Figure~\ref{fig:fig1}, the model exhibits two 
distinct behavioral regimes. 
For problems within its capability 
boundary, it engages in productive reasoning and derives correct 
solutions. However, when operating beyond this boundary, it fails 
to recognize its limitations and instead produces spurious outputs. 
For instance, on a Countdown task requiring construction of target 0.3 from numbers [1,5,6,8,9], the model attempts multiple approaches but ultimately fabricates an incorrect expression and falsely claims it evaluates to 0.3 (the actual result is 0.6).
Instead, it 
should simply acknowledge ``I cannot reach 0.3 with the given numbers''.
Such failures typically manifest as specious logical steps or endless generation, creating a deceptive illusion of competence.
Ideally, models should attempt to reason on challenging problems, 
but abort and explicitly refuse when they recognize, after several 
reasoning steps, that a solution is unattainable.

% To quantify this phenomenon, we conduct a comprehensive empirical 
% study using a difficulty-graded Countdown task ($N=3$ to $N=8$). 
% We observe that models exhibit universal capability overreach: 
% vanilla models never refuse (0\% across all difficulty levels), 
% and even explicit prompting fails to prevent over 80\% of futile 
% attempts on the hardest tasks.
% % 
% The dominant failure mode is 
% \textit{specious reasoning}—superficially valid outputs containing 
% subtle hallucinations—whose prevalence escalates with task difficulty. 
% Moreover, models are systematically miscalibrated, exhibiting 
% over-confidence (20\%) 6× more frequently than over-conservatism (3.4\%).
% % 
% Our analysis reveals that prompt engineering alone is insufficient: it reduces Refusal Recall below 30\% on hard tasks while causing 10\% Capability Loss on easy ones, necessitating a learning-based approach to enforce behavioral alignment.

To quantify this phenomenon, we conduct a comprehensive empirical 
study using a difficulty-graded Countdown task ($N=3$ to $N=8$). 
We observe that models exhibit universal capability overreach: 
vanilla models never refuse (0\% across all difficulty levels), 
and even explicit prompting fails to prevent over 80\% of futile 
attempts on the hardest tasks. The dominant failure mode is 
\textit{specious reasoning}, superficially valid outputs containing 
subtle hallucinations, whose prevalence escalates with task difficulty.
Moreover, by quantifying the alignment gap under standard prompting, we reveal systematic miscalibration with over-confidence (20\%) occurring 6× more frequently than over-conservatism (3.4\%), and reducing refusal recall below 30\% on hard tasks while causing 10\% capability loss on easy ones, indicating that stronger interventions are required.

Building on these insights, we introduce \textbf{\textsc{CaRL}} 
(\textbf{Ca}pability-\textbf{a}ligned \textbf{R}einforcement 
\textbf{L}earning), a framework designed to align model behavior 
with capability boundaries. CaRL operates through two mechanisms: 
(1) \textbf{Capability-Calibrated Reward Shaping}, which establishes 
a strict reward hierarchy to incentivize refusal 
over futile reasoning, and (2) \textbf{Hindsight Refusal Augmentation}, 
which enhances refusal samples by converting futile reasoning into 
refusal format.
Together, these mechanisms enable models to explore solution paths 
and abort when they detect insurmountable difficulties, rather than 
fabricating specious outputs. CaRL can be applied to an arbitrary 
model to instill capability boundary awareness. Experiments on Qwen3-8B and Qwen3-14B
demonstrate that CaRL preserves performance on both in-distribution 
and out-of-distribution tasks while reducing futile reasoning (8B: 65.5\% → 7.0\%; 14B: 78.6\% → 1.0\%), effectively calibrating the model's 
behavioral boundary with its true competence.

Our contributions are summarized as follows:

\begin{itemize}
    \item We identify and characterize \textbf{\textit{futile reasoning}},
    demonstrating that models generate plausible-looking but fundamentally incorrect reasoning when facing beyond-capability tasks.
    
    \item We conduct a systematic empirical analysis, and reveal three critical 
    phenomena: (1) universal capability overreach, 
    (2) specious reasoning as the dominant failure mode escalating with 
    difficulty, and (3) a substantial capability-behavior alignment gap with 6× bias towards over-confidence, demonstrating that implicit behavioral cues are insufficient.
    \item We propose \textbf{\textsc{CaRL}}, combining capability-calibrated 
    reward shaping and hindsight refusal augmentation, achieving substantial reduction in futile reasoning while preserving task performance.
\end{itemize}
% \section{Related Work}
% Recently, \citet{chen2024not} has shown that llm 大多数错误情况是由于 overthinking 导致的。 \cite{yeo2025demystifying} 表明长思考模型的一大困难就是模型复读。这些indicate repetition and specious maybe futile reasoning 中的典型现象。
% % 
% 对于 overconfidence 问题，\citet{cheng2024can,kadavath2022language} 的文章探究了模型的拒答能力以及自我知识认知，然而它们主要关注知识边界，而忽略了模型的能力边界。
% 近期，\cite{joglekar2025training} 也表明模型的honest via 忏悔能力非常重要

\section{Related Work}

\paragraph{Futile Reasoning.}
Recent analysis has identified specific pathologies in model reasoning. 
\citet{chen2024not} demonstrates that a significant proportion of errors in LLMs stems from overthinking.
Furthermore, \citet{yeo2025demystifying} highlights that a primary obstacle for long-thinking models is the tendency towards repetition.
These findings corroborate our characterization of \textit{futile reasoning}, where models engage in repetitive loops or generate specious justifications that appear plausible but are functionally void.

\paragraph{Capability Alignment and Honesty.}
To mitigate overconfidence, prior works have explored the model's ability to refuse and its self-knowledge~\cite{cheng2024can, kadavath2022language}. 
However, these studies predominantly focus on \textit{knowledge boundaries} (i.e., factual retrieval), largely overlooking the \textit{capability boundaries} essential for reasoning tasks.
More recently, \citet{joglekar2025training} emphasizes the importance of achieving honesty via confession, suggesting that models should be trained to recognize and correct their own errors.
Our work extends this by using reinforcement learning to internalize this repentance, teaching models to abort futile reasoning when intrinsic capability limits are reached.
% \section{Preliminary Experiments and Analyses}

% To systematically understand the disconnect between a model's confidence and its actual capabilities, we conducted a series of diagnostic experiments. 

% These experiments were designed to answer three key questions: 
% 1. 模型是否会承认自己的不知道，or capability-overreach
% 2. 错误案例是什么样的
% 3. 对能力边界的认知是否和模型真实能力具有一致性

% \subsection{Experimental Setup: The Countdown Task}

\section{Preliminary Experiments and Analyses}

We conduct a systematic empirical analysis to characterize the nature 
of \textit{futile reasoning}. 
Our primary objective is to quantify the 
misalignment between LLMs' generative behavior and their intrinsic 
competence boundaries. 
% Using a controlled reasoning testbed, we design 
% our experiments to investigate three fundamental aspects: 
% (1) \textbf{Behavioral Tendency:} Does the model acknowledge its 
% limitations when tasks exceed its capability, or does it exhibit 
% capability overreach?
% (2) \textbf{Failure Characterization:} What are the characteristic 
% failure patterns when the model operates beyond its capability boundary?
% (3) \textbf{Alignment Gap:} Is the model's self-assessed capability 
% boundary consistent with its empirical performance?
% 
Using a controlled reasoning testbed, we investigate three key questions: 
(1) Do models refuse beyond-capability tasks, or exhibit universal 
capability overreach?
(2) What are the characteristic failure patterns of futile 
reasoning?
(3) Are models' refusal decisions properly calibrated with their empirical capabilities?

\subsection{Experimental Setup: The Countdown Task}

To decouple \textit{reasoning capability} from knowledge retrieval~\cite{huang2025survey,sahoo2024comprehensive}, 
we utilize the Countdown Task (a variant of the Game of 24) as our primary 
testbed, as it allows precise manipulation of difficulty without confounding 
factors of external knowledge. 
A detailed example is shown in Appendix~\ref{countdown}.

% Given a set of $N$ input numbers, the model 
% must use basic arithmetic operations ($+, -, \times, \div$) to reach a target 
% value, with each number usable at most once. We control difficulty by varying 
% $N$ from 3 (Easy) to 8 (Hard), making the problem combinatorially more complex.

% % 对应问题 1: 模型是否会承认自己的不知道

\subsection{Phenomenon I: Universal Capability Overreach}
\label{sec:phenomenon1}

\begin{figure}[H]
    \centering
    \includegraphics[width=1.0\linewidth]{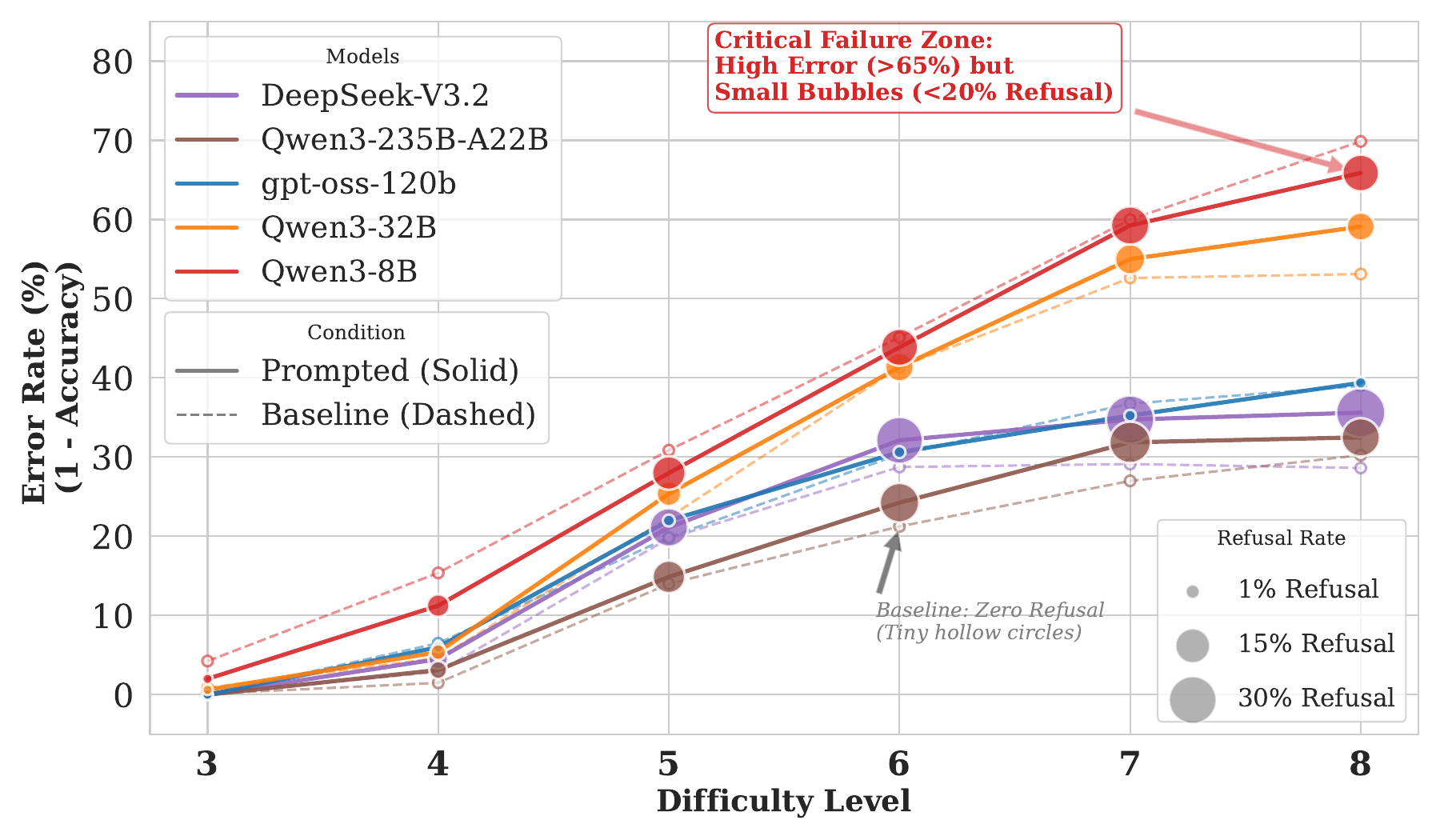}
    % \caption{Capability Overreach in Qwen-32B.}
    \caption{Universal Capability Overreach.}
    \label{fig:phenomenon_1}
\end{figure}
We first investigate whether models possess an intrinsic mechanism to acknowledge ignorance when facing unsolvable problems. 
We conduct experiments on  Qwen3-8B~\cite{yang2025qwen3}, Qwen3-32B~\cite{yang2025qwen3}, gpt-oss-120b~\cite{openai2025gptoss120bgptoss20bmodel}, Qwen3-235B-A22B~\cite{yang2025qwen3}, and DeepSeek-V3.2~\cite{liu2025deepseek} across difficulty levels ($N=3$ to $N=8$). 
We experiment on two settings: ``Baseline'' refers to standard generation with long chain-of-thought reasoning, while ``Prompted'' employs explicit instructions encouraging the model to acknowledge ignorance when facing beyond-capability tasks.

As illustrated in Figure~\ref{fig:phenomenon_1}, we report the error rate of each model while visualizing its refusal rate through the radius of each data point. 
Ideally, a well-calibrated model should exhibit a dynamic response: as the error rate increases, the radius of the point should expand significantly, indicating that the model correctly identifies the rising difficulty and opts to refuse rather than hallucinate.

However, we observe a critical calibration failure. 
Despite error rates surging with task difficulty (e.g., Qwen3-8B exceeds 65\% error at $N=8$), the corresponding bubbles remain negligibly small across all models. 
Even in the prompted setting, the marginal expansion in bubble size is disproportionate to the massive increase in errors. The persistence of these tiny bubbles at high error rates reveals that models blindly attempt to reason through unsolvable queries despite explicit instructions to the contrary.

Specifically, gpt-oss-120b remains obstinately non-compliant, failing to refuse even under the prompted setting. 
In contrast, Qwen3-235B-A22B and DeepSeek-V3.2 demonstrate latent refusal capabilities that can be effectively elicited through prompting.
Moreover, DeepSeek-V3.2 is the unique model that displays traces of spontaneous self-doubt (albeit in <1\%) under the baseline setting. 
\textbf{All these phenomena show that models lack intrinsic mechanisms to recognize capability boundaries, exhibiting universal capability overreach regardless of model scale or task difficulty.
}

\subsection{Phenomenon II: Failure Patterns of Futile Reasoning}
\label{sec:anatomy}

\begin{figure}[H]
    \centering
    \includegraphics[width=1.0\linewidth]{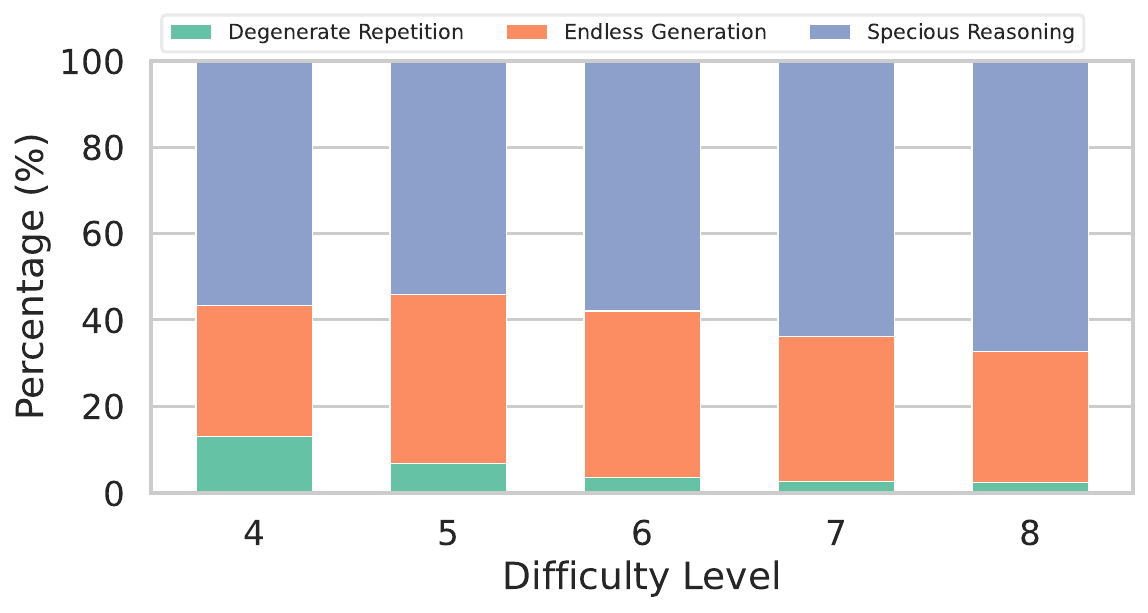}
    \caption{Distribution of Futile Reasoning Patterns.}
    \label{fig:futile_dist}
\end{figure}
We categorize futile reasoning into three distinct patterns. \textit{Specious 
Reasoning} constructs apparently coherent derivations that contain subtle 
errors (e.g., arithmetic mistakes, reusing numbers), creating an illusion 
of competence. \textit{Endless Generation} continuously proposes new attempts 
without recognizing futility, exhaustively exploring combinations. 
\textit{Degenerate Repetition} enters recursive loops, repeating identical 
reasoning steps.

Figure~\ref{fig:futile_dist} reveals that specious reasoning dominates 
(57-68\% across levels), while endless generation remains stable (30-40\%) 
and degenerate repetition decreases (13\%→2\%). Critically, as difficulty 
increases, models shift from simple loops to suspicious reasoning.This indicates that 
\textbf{when pushed beyond capability, models do not degenerate randomly; 
instead, they fabricate increasingly sophisticated justifications to maintain a facade of correctness}.

\subsection{Phenomenon III: Quantifying Capability-Behavior Misalignment}
\label{sec:miscalibration}

We quantify the alignment gap between ideal capability-aware behavior 
and actual model responses. 
Following prior work~\cite{lin2022teaching,kadavath2022language,cheng2024can}, 
we sample multiple responses per question and classify it as solvable
if the model achieves an accuracy > 0.5, and unsolvable otherwise.
We then define the \textbf{Capability Quadrants}, categorizing model behaviors 
into four outcomes: A. \textit{Ideal Answer} (solvable, answered), 
B. \textit{Over-Conservative} (solvable, refused), C. \textit{Over-Confidence} 
(unsolvable, answered), and D. \textit{Ideal Refusal} (unsolvable, refused).
We analyze Qwen3-32B responses in this section

\begin{figure}[H]
    \centering
    \includegraphics[width=0.7\linewidth]{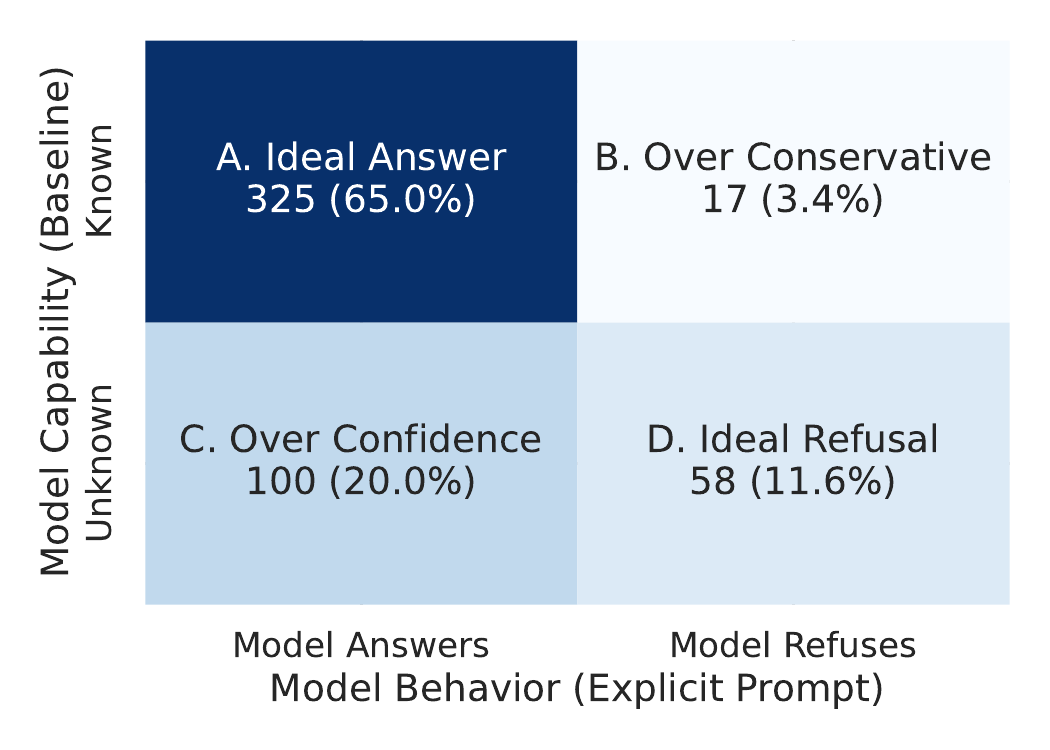}
    \caption{Distribution of Capability Quadrants.}
    \label{fig:distribution}
\end{figure}

\paragraph{Asymmetry of Misalignment.}
Figure~\ref{fig:distribution} reveals a striking 6× asymmetry: Over-Confidence 
(20\%) vastly exceeds Over-Conservative (3.4\%). This directional bias indicates 
\textbf{models systematically over-estimate capability rather than exhibiting random uncertainty}.

\paragraph{Degradation Across the Difficulty Spectrum.}

\begin{figure}[H]
    \centering
    \includegraphics[width=1.0\linewidth]{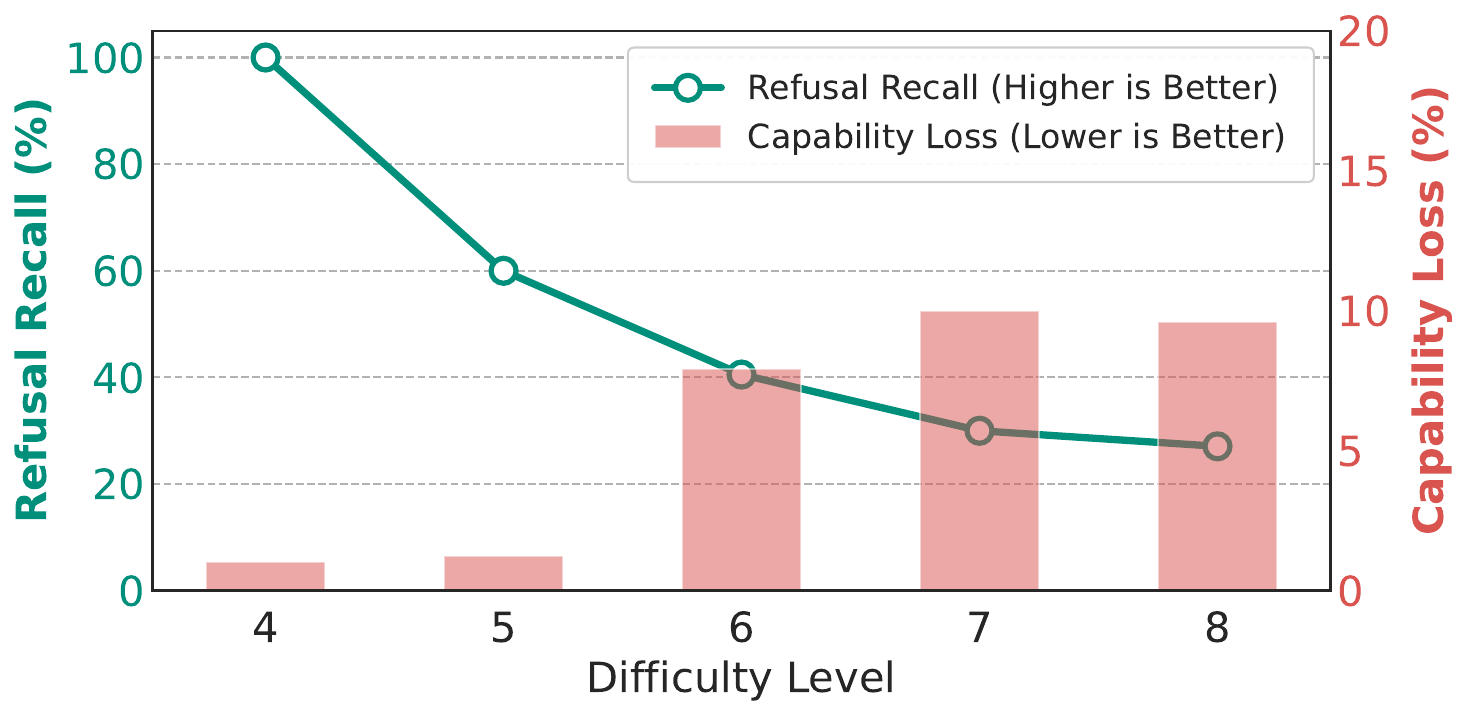}
    \caption{\textbf{The Alignment Trade-off.} Naive prompting leads to a collapse in Refusal Recall (Green) on hard tasks while simultaneously increasing Capability Loss (Pink) on solvable tasks.}
    \label{fig:tradeoff}
\end{figure}

To understand how the capability quadrants shift with task difficulty, 
we examine two aggregate metrics derived from the quadrant distribution. 
\textit{Refusal Recall} measures safety by quantifying what fraction of 
unsolvable tasks fall into the \textit{Ideal Refusal} quadrant (versus 
\textit{Over-Confidence}). \textit{Capability Loss} measures utility by 
quantifying what fraction of solvable tasks fall into the \textit{Over-Conservative} 
quadrant (versus \textit{Ideal Answer}).
Figure~\ref{fig:tradeoff} shows that as difficulty increases, Refusal Recall 
collapses (100\%→30\%) while Capability Loss rises (0\%→10\%). This reveals 
\textbf{poor calibration: models apply uniform shifts that underprotect hard 
tasks while harming solvable ones}.

% Figure~\ref{fig:tradeoff} visualizes these metrics across difficulty levels. 
% Refusal Recall degrades dramatically from near 100\% at $N=3$ to below 30\% 
% at $N=8$, indicating that the \textit{Over-Confidence} quadrant expands 
% while \textit{Ideal Refusal} shrinks as tasks become harder. Simultaneously, 
% Capability Loss gradually increases from 0\% to 10\% as difficulty scales, 
% showing that \textit{Over-Conservatism} becomes more severe. This reveals 
% a poorly calibrated adjustment mechanism: \textbf{models respond to increasing 
% difficulty by becoming more conservative overall, yet this adjustment is 
% insufficient for truly unsolvable tasks (Refusal Recall collapses to 30\%) 
% while simultaneously harming solvable ones (10\% incorrectly refused). 
% The model cannot distinguish task difficulty granularly, applying a coarse, 
% uniform shift that underprotects safety while degrading utility}.

\begin{figure}[H]
    \centering
    \includegraphics[width=1.0\linewidth]{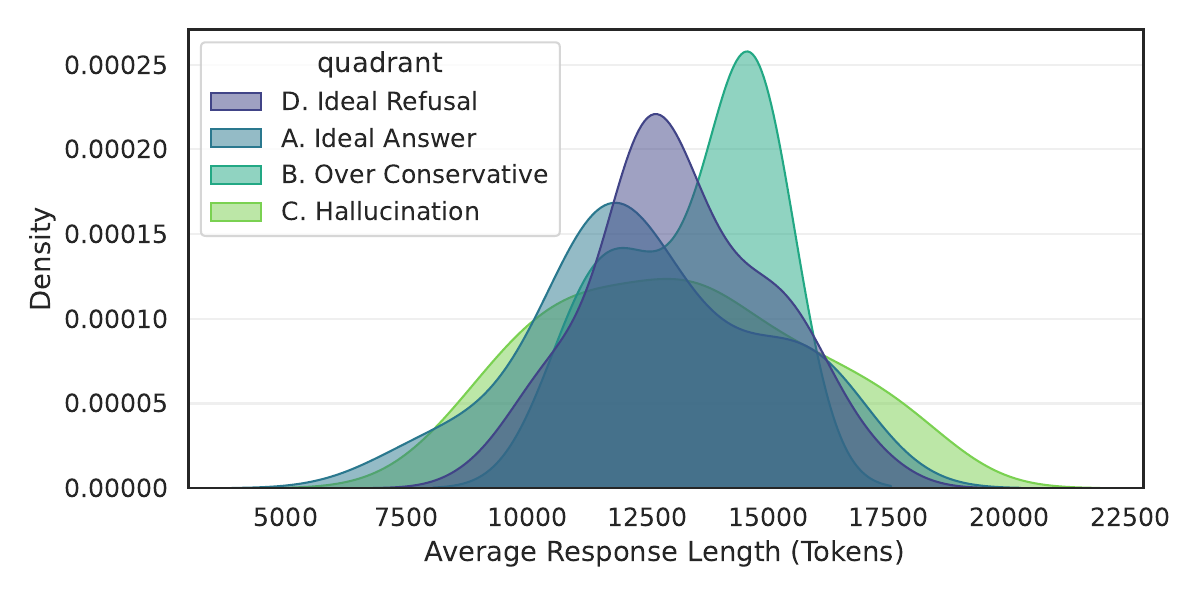}
    \caption{\textbf{Reasoning Depth Distribution.} Refusal behaviors show decisive termination (Peaks), whereas Over-Confidence exhibits a long-tail distribution, confirming the high computational cost of futile reasoning.}
    \label{fig:len_dist}
\end{figure}

\paragraph{Computational Cost of Misalignment.}
Beyond correctness, misalignment imposes computational costs. 
Figure~\ref{fig:len_dist} shows token length distributions across 
behavioral quadrants based on KDE~\cite{davis2011remarks,parzen1962estimation}. Refusal behaviors (Ideal Refusal and Over-Conservative) exhibit sharply peaked distributions, indicating decisive termination. 
In contrast, Over-Confidence shows a flat distribution with a pronounced 
long tail, generating 2-3× more tokens than proper refusals. This confirms 
that when models attempt unsolvable tasks, they engage in prolonged futile 
reasoning, consuming excessive computation without producing value.
Consequently, mitigating the misalignment may offer an additional benefit 
beyond correctness: reducing unnecessary computational overhead from 
futile reasoning attempts.

\begin{figure*}
    \centering
    \includegraphics[width=1.0\linewidth]{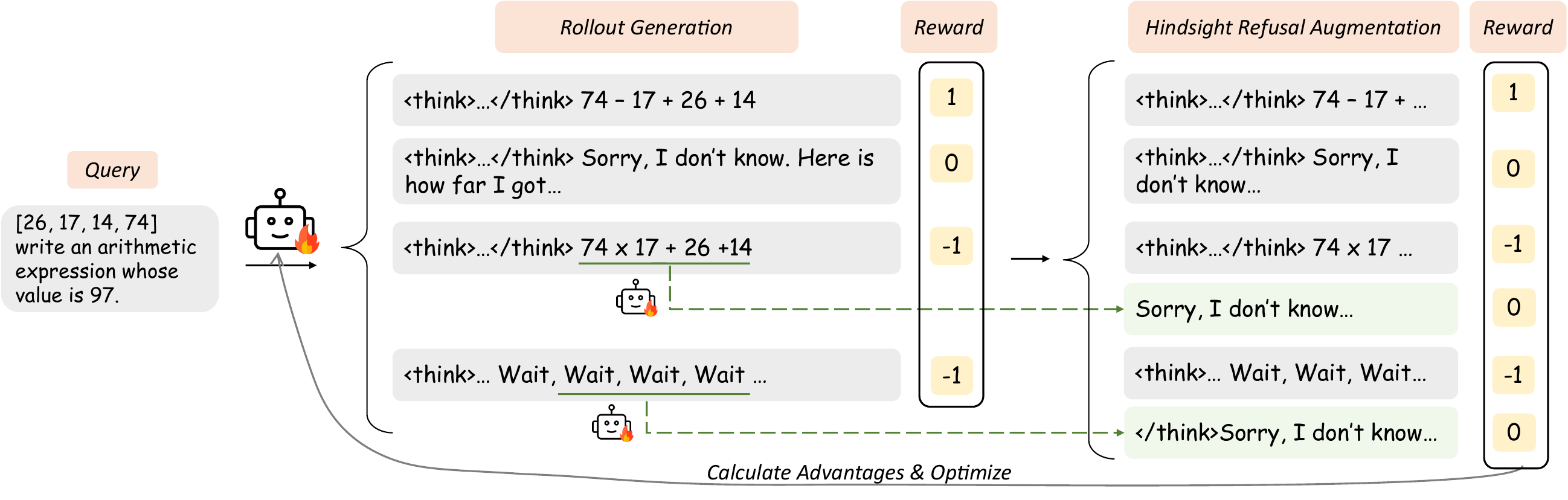}
    \caption{\textbf{Overview of CaRL.} Our framework combines 
(1) \textbf{Capability-Calibrated Reward Shaping} that establishes a 
preference hierarchy favoring refusal over hallucination, and 
(2) \textbf{Hindsight Refusal Augmentation} that converts failed trajectories 
into refusal trajectories, enabling capability-aligned behavior.}
    \label{fig:method}
\end{figure*}

\section{Capability-aligned Reinforcement Learning}

% We propose \textbf{CaRL} (\textbf{Ca}pability-\textbf{a}ligned \textbf{R}einforcement \textbf{L}earning), a strictly constrained RL framework designed to penalize capability overreach. 
% Rather than introducing complex architectural changes, CaRL aligns the model's behavioral boundary with its intrinsic capability through two targeted mechanisms:

% \begin{enumerate}
%     \item \textbf{Capability-Calibrated Reward Shaping}, which establishes a strict preference hierarchy ($+1 > 0 > -1$) to incentivize safety over specious gambling.
%     \item \textbf{Hindsight Refusal Augmentation (HRA)}, a data-centric strategy that enriches the training buffer with valid refusal supervision derived from capability failures.
% \end{enumerate}

To alleviate futile reasoning, we propose \textbf{CaRL} (\textbf{Ca}pability-\textbf{a}ligned 
\textbf{R}einforcement \textbf{L}earning), a framework that aligns model 
behavior with capability boundaries through explicit reinforcement learning. 
As shown in Figure~\ref{fig:method}, 
CaRL consists of two main components:
(1) \textbf{Capability-Calibrated Reward Shaping} (§\ref{sec:reward}) encodes capability boundaries into the reward structure through a hierarchical design that penalizes capability overreach while preserving the incentive for productive reasoning.
(2) \textbf{Hindsight Refusal Augmentation} (§\ref{sec:hra}) addresses the scarcity of refusal training data by retrospectively converting failed reasoning attempts into refusal demonstrations, densifying the sparse reward landscape for refusal.

\subsection{Capability-Calibrated Reward Shaping}
\label{sec:reward}
Standard reasoning RL typically assigns 1 for correct answers and 0 for incorrect answers. This leads to uniform low rewards to 
both incorrect answers and refusals, treating them as equivalent failures. 
This creates no incentive for the model to distinguish between attempting 
and refusing, encouraging futile reasoning since there is no penalty for 
generating plausible-sounding but incorrect outputs.

To rectify this, we reshape the reward function to 
explicitly encode the preference for safe refusal over futile attempts. We define the reward $r(c)$ for a generated solution $c$ as:
\begin{equation}
    r(c) = 
    \begin{cases} 
    +1 & \text{if } c \text{ is correct} \\
    \phantom{+}0 & \text{if } c \text{ is a valid refusal} \\
    -1 & \text{if } c \text{ is incorrect} 
    \end{cases}
    \label{eq:reward}
\end{equation}
where a valid refusal is identified by explicit refusal snippets (e.g., 
``Sorry, I can't solve the problem.'').

% By positioning refusal ($r=0$) strictly between success ($r=+1$) and 
% failure ($r=-1$), we create a value gradient that incentivizes the policy 
% to transition towards the refusal state when the expected success probability 
% is low. Specifically, under the GRPO objective:
% \begin{equation}
%     \mathcal{L}_{\text{GRPO}} = \mathbb{E}_{\tau \sim \pi_\theta} 
%     \left[ (r(\tau) - \bar{r}) \log \pi_\theta(\tau) \right]
% \end{equation}
% where $\bar{r}$ is the group-wise baseline reward, the policy learns to 
% increase the probability of refusal when it yields higher rewards than 
% incorrect attempts. This design directly counteracts the 6× over-confidence 
% bias quantified in §\ref{sec:miscalibration}, establishing a safety margin 
% that penalizes capability overreach.

\subsection{Hindsight Refusal Augmentation}
\label{sec:hra}

While reward shaping provides the incentive for refusal, it does 
not solve the data scarcity problem. Models fine-tuned for 
helpfulness rarely generate refusals during on-policy exploration~\cite{kashyap2025too,huang2025position,bai2022training}, and RL exploration is notoriously inefficient in unfamiliar behavioral regions~\cite{chen2025beyond,chen2025sft,zhang2025policy}. 
Our analysis confirms this challenge: as shown in §\ref{sec:miscalibration}, baseline models exhibit 0\% refusal rate, indicating that appropriate refusal lies far outside the model's natural behavioral distribution. 
Consequently, on-policy exploration would require prohibitively many samples before encountering sufficient refusal examples, rendering the reward structure alone insufficient for effective learning.

We address this through \textbf{Hindsight Refusal Augmentation (HRA)}. 
Rather than waiting for rare on-policy refusals, we leverage a key observation: 
\textit{every failed reasoning attempt implicitly reveals a situation where 
refusal would have been the appropriate action}. By retrospectively converting 
these abundant failures into synthetic refusal demonstrations, HRA densifies 
the sparse $r=0$ training signal, enabling the policy to learn refusal 
boundaries efficiently.

For each on-policy trajectory $\tau = (q, c)$ that generates an incorrect 
answer $c$ (receiving $r=-1$) based on a query $q$, we construct an augmented refusal sample 
$\tau_{\text{aug}} = (q, c_{\text{refuse}})$ as follows. We first retain 
the reasoning trace up to (but excluding) the final answer step, then insert 
a refusal prefix such as ``Sorry, I cannot solve this problem. Here is how 
far I got:''. Next, we prompt the model to generate a brief summary of the 
attempted reasoning progress, creating a natural refusal response that explains 
what was tried. Finally, we assign the refusal reward $r=0$ to this modified 
trajectory and add it to the training batch alongside the original failure.

This creates a contrastive learning signal: under the same reasoning context 
that led to failure, the model learns that choosing to refuse ($r=0$) yields 
higher reward than persisting to an incorrect conclusion ($r=-1$). By 
systematically augmenting every failure with its refusal counterpart, HRA 
densifies the sparse reward landscape, enabling the policy to learn appropriate 
refusal boundaries even when on-policy exploration rarely produces refusals.

\subsection{Training Procedure}

Algorithm~\ref{alg:carl} details the complete training procedure. At each iteration, 
we generate multiple outputs per question, compute capability-calibrated 
rewards (§\ref{sec:reward}), and apply HRA to augment failed attempts with 
refusal samples (§\ref{sec:hra}). The policy is then updated using group-relative advantages computed from the augmented dataset. We employ the GRPO~\cite{liu2024deepseek,guo2025deepseek} objective for optimization.

\begin{algorithm}[H]
\caption{CaRL Training Procedure}
\footnotesize
\label{alg:carl}
\begin{algorithmic}[1]
\Statex \hspace{-\algorithmicindent} \textbf{Input:} Policy $\pi_{\theta}$, training set $\mathcal{Q}$, group size $G$, batch size $B$
\For{iteration $t = 1, \ldots, T$}
    \State Sample batch: $\{q_i\}_{i=1}^{B} \sim \mathcal{Q}$
    \State \textcolor{blue}{// Rollout Generation}
    \State Generate $G$ outputs per question: $\{c_{i,j}\}_{j=1}^{G} \sim \pi_{\theta}$ for each $q_i$
    \State \textcolor{blue}{// Capability-Calibrated Reward Shaping}
    \State Compute rewards via Eq.~\eqref{eq:reward}: $r_{i,j} \leftarrow r(c_{i,j}$)
    \State Collect rollouts: $\mathcal{D}_t \leftarrow \{(\tau_{i,j}, r_{i,j})\}$ where $\tau_{i,j} = (q_i, c_{i,j})$
    \State \textcolor{blue}{// Hindsight Refusal Augmentation}
    \For{each $(\tau_{i,j}, r_{i,j}) \in \mathcal{D}_t$ where $r_{i,j} = -1$}
        \State Construct refusal rollout: $\tau_{i,j}^{\text{aug}}$
        \State Augment: $\mathcal{D}_t \leftarrow \mathcal{D}_t \cup \{(\tau_{i,j}^{\text{aug}}, r=0)\}$
    \EndFor
    \State \textcolor{blue}{// Policy update via GRPO}
    \State Compute group advantages and update $\pi_{\theta}$
\EndFor
\Statex \hspace{-\algorithmicindent} \textbf{Output:} $\pi_{\theta}$
\end{algorithmic}
\end{algorithm}

\section{Experiments}

\begin{table*}[t]
    \centering
    \resizebox{\linewidth}{!}{
    \newcommand{\good}[1]{\textcolor{green!40!black}{\tiny~(#1)}}
    \newcommand{\bad}[1]{\textcolor{red!60!black}{\tiny~(#1)}}

    \begin{tabular}{l|llcl|llcl}
        \toprule
        \multirow{2}{*}{\textbf{Method}} & \multicolumn{4}{c|}{\textbf{In-Distribution (Countdown)}} & \multicolumn{4}{c}{\textbf{Out-of-Distribution (Sudoku)}} \\
        \cmidrule(lr){2-5} \cmidrule(lr){6-9}
         & \textbf{Acc} $\uparrow$ & \textbf{Reliability} $\uparrow$ & \textbf{Refusal} & \textbf{Futile} $\downarrow$ & \textbf{Acc} $\uparrow$ & \textbf{Reliability} $\uparrow$ & \textbf{Refusal} & \textbf{Futile} $\downarrow$ \\
        \midrule
        \multicolumn{9}{l}{\textit{Qwen3-8B}} \\
        \midrule
        Vanilla & 59.67 & 0.6663 & 13.92 & 65.50 & 46.88 & 0.4969 & 10.62 & 89.41 \\
        Standard RL & 64.08\good{+4.4} & 0.6425\bad{-.02} & 0.33 & 99.00\bad{+33.5} & 43.25\bad{-3.6} & 0.4744\bad{-.02} & 13.13 & 85.00\good{-4.4} \\
        RL\textsubscript{unk=0} & 62.71\good{+3.0} & 0.6296\bad{-.04} & 0.50 & 99.00\bad{+33.5} & 44.62\bad{-2.3} & 0.4850\bad{-.01} & 12.12 & 86.00\good{-3.4} \\
        RL\textsubscript{unk=0.5} & 63.42\good{+3.8} & 0.6371\bad{-.03} & 0.58 & 98.00\bad{+32.5} & 45.25\bad{-1.6} & 0.5131\good{+.02} & 16.75 & 78.00\good{-11.4} \\
        RFT & 59.13\bad{-0.5} & 0.7610\good{+.09} & 35.21 & 17.00\good{-48.5} & 0.00\bad{-46.9} & 0.4763\bad{-.02} & 95.25 & 5.00$^\dagger$ \\
        \textbf{CaRL (Ours)} & 61.00\good{+1.3} & \textbf{0.7915}\good{+.13} & 37.17 & \textbf{7.00}\good{-58.5} & 46.25\bad{-0.6} & \textbf{0.6156}\good{+.12} & 36.63 & \textbf{43.02}\good{-46.4} \\
        \midrule
        \multicolumn{9}{l}{\textit{Qwen3-14B}} \\
        \midrule
        Vanilla & 63.25 & 0.6719 & 7.88 & 78.57 & 50.75 & 0.5556 & 18.62 & 80.46 \\
        Standard RL & 56.42\bad{-6.8} & 0.5750\bad{-.10} & 2.17 & 95.03\bad{+16.5} & 43.63\bad{-7.1} & 0.4831\bad{-.07} & 13.13 & 83.00\bad{+2.5} \\
        RL\textsubscript{unk=0} & 68.21\good{+5.0} & 0.8050\good{+.13} & 24.58 & 23.00\good{-55.6} & 48.38\bad{-2.4} & 0.5369\bad{-.02} & 14.00 & 79.00\good{-1.5} \\
        RFT & 63.12\bad{-0.1} & 0.7879\good{+.12} & 31.33 & 15.00\good{-63.6} & 0.00\bad{-50.8} & 0.4525\bad{-.10} & 90.50 & 10.00$^\dagger$ \\
        \textbf{CaRL (Ours)} & 67.25\good{+4.0} & \textbf{0.8348}\good{+.16} & 32.50 & \textbf{1.00}\good{-77.6} & 44.87\bad{-5.9} & \textbf{0.6262}\good{+.07} & 38.87 & \textbf{36.00}\good{-44.5} \\
        \bottomrule
        \multicolumn{9}{l}{\footnotesize $^\dagger$ RFT's low futile rate on OOD is a trivial result of collapsing into near-total refusal (Ref $>90\%$, Acc $=0\%$).}
    \end{tabular}}
    \caption{\textbf{Main Results on In-Distribution and Out-of-Distribution Tasks.} 
    Values in parentheses show changes relative to Vanilla baseline. 
    \textcolor{green!40!black}{Green} indicates improvement, 
    \textcolor{red!60!black}{red} indicates degradation.}
    \label{tab:combined_results}
\end{table*}

\subsection{Datasets}

% Our experimental datasets are constructed using the \texttt{reasoning-gym} framework\footnote{\url{https://github.com/open-thought/reasoning-gym}}. 
% For the primary capability assessment, we utilize the \textbf{Countdown} task, synthesizing data across three distinct difficulty levels ($N=4, 6, 8$). 
% For each difficulty tier, we generate a training set of 1,000 samples and a held-out evaluation set of 100 samples, ensuring a balanced distribution of complexity. 
% Additionally, to evaluate the method's generalization capability on out-of-distribution (OOD) tasks, we synthesize a separate \textbf{Sudoku} dataset using the same framework with 100 samples. Specially, Sudoku 通常相比 countdown 需要更多的求解步骤，可以较好的反应模型在面对困难问题时是否有过度拒答的情况

We utilize the \texttt{reasoning-gym} framework\footnote{\url{https://github.com/open-thought/reasoning-gym}} to construct our benchmarks. 
For the in-distribution \textbf{Countdown} task, we generate 1,000 training and 100 evaluation samples for each of the three difficulty levels ($N=4, 6, 8$). 
For out-of-distribution (OOD) evaluation, we synthesize 100 \textbf{Sudoku} samples, which require filling a partially completed 9×9 grid such that each row, column, and 3×3 subgrid contains all digits 1-9 without repetition.
Notably, Sudoku puzzles necessitate significantly longer reasoning chains compared to Countdown tasks. We leverage this characteristic as a stress test to verify whether the model maintains robust reasoning on complex queries, or if it simply collapses into over-refusal when facing high computational costs.

\subsection{Baselines}

We evaluate performance against the following baselines. 
\textbf{Vanilla} refers to the raw reasoning model without further alignment. 
\textbf{Standard RL} optimizes solely for task accuracy using a binary reward structure ($r=1$ for correct answers, $r=-1$ otherwise). 
RFT~\cite{zhang2023cumulative} (Rejection Sampling Fine-Tuning) performs supervised fine-tuning on a static dataset constructed by filtering for both correct reasoning paths and valid refusals. 
RL\textsubscript{unk} adopts our capability-calibrated reward shaping to decouple the impact of reward signals from HRA. Specifically, we implement two strategies: \textbf{RL\textsubscript{unk=0}} assigns a neutral reward ($r=0$) to refusal responses, while \textbf{RL\textsubscript{unk=0.5}} assigns a positive partial reward ($r=0.5$).

% \subsection{Evaluation Metrics}
% To comprehensively report model's performance of futile reasoning, we report four metrics:
% \textbf{Acc} refers to task accuracy.
% \textbf{Reliability Score} refers to 综合分数，1 分 for 正确，0.5 unknown，0 分错误。
% Moreover，我们汇报了两个指标给，\textbf{Refusal Rate} 统计模型拒答的比例，and \textbf{Futile Rate} 统计模型回答错误的里面没有拒答的比例。

% \subsection{Evaluation Metrics}

% To comprehensively assess the model's performance and alignment boundaries, we utilize four quantitative metrics. 
% First, we report \textbf{Acc} (Accuracy) to measure the standard success rate on the reasoning tasks. 
% To evaluate the trade-off between correctness and safety, we introduce the \textbf{Reliability Score}, a weighted metric that assigns a value of $1.0$ for correct solutions, $0.5$ for valid refusals (``Unknown''), and $0$ for incorrect answers. 
% Furthermore, to characterize the model's behavioral tendencies, we report the \textbf{Refusal Rate}, which measures the overall proportion of queries where the model admits ignorance. 
% Finally, we calculate the \textbf{Futile Rate}, defined as the proportion of incorrect responses where the model failed to refuse (i.e., hallucinations), quantifying the extent of capability overreach within failure cases.

\subsection{Evaluation Metrics}

To comprehensively assess the model's performance and alignment boundaries, 
we utilize four quantitative metrics. Let $N_c$, $N_r$, and $N_i$ denote the 
number of correct, refusal, and incorrect responses, respectively, with 
$N = N_c + N_r + N_i$ as the total number of samples.

\paragraph{Accuracy (Acc).} Success rate: $\text{Acc} = N_c/N$.

\paragraph{Reliability Score.} 
Weighted metric assigning 1.0 for correct, 0.5 for refusal, 0 for incorrect: 
$\text{Reliability} = (N_c + 0.5 N_r) / N$.

\paragraph{Refusal Rate.} 
Proportion of acknowledged ignorance: $\text{Refusal Rate} = N_r / N$.

\paragraph{Futile Rate.} 
Proportion of failures where the model generates futile reasoning instead of refusing: 
$\text{Futile Rate} = N_i / (N_i + N_r)$.

% \subsection{Implementation details}
% We employ GRPO as our RL alogthrm.
% We training 3 epoch with learning rate 8e-5 in rft, train batch size 32.
% We training 5 epoch  with learning rate 1e-6 in rl and rollout n=16, train batch size 32, temperature 1.0.

% For all evaluation, we sample 8 取平均 with temperature 0.6.

\subsection{Main results}
% As shown in Table~\ref{tab:combined_results}, we report the evaluation result of Qwen3-8B. We have the following findings:

% \textbf{CaRL} achieves good performance on both in-distribution and out-of-distribution settings. It can 在维持性能不降低的情况下，保持良好的拒答率。

% \textbf{Reply on simple reward can't 激发 模型的拒答能力} RL 方法在 futile 上几乎没降，这表明在rl 过程中模型倾向于自信的进行探索而不是学习拒答能力。特别地，RL unk=0 和 RL unk=0.5 的性能没有较大差异，佐证了这一点。

% \textbf{RFT 无法激发模型的泛化性} 尽管相比于 RL baseline， RFT 在 in-distribution 表现出了较好的，然后它在 ood 上呈现出了过度的拒答，而难以尝试得到正确的分数。

As presented in Table~\ref{tab:combined_results}, we evaluate Qwen3-8B and Qwen3-14B across in-distribution (Countdown) and out-of-distribution (Sudoku) settings, comparing against the vanilla baseline. The analysis reveals three critical findings. 

% \paragraph{CaRL achieves optimal performance across scales.} 
% CaRL consistently attains the highest Reliability and lowest futile rate across 
% both model sizes. On Qwen3-8B in-distribution, CaRL achieves Reliability 0.7915 (+0.13) with 7.00\% futile rate (-58.5 points). 
% On 14B, these improvements amplify: Reliability 0.8348 (+0.16) with merely 1.00\% futile rate (-77.6 points), demonstrating that larger models benefit more from capability-aligned training. 
% % 
% In contrast, Standard RL improves accuracy on 7B (+4.4) but 
% catastrophically increases futile reasoning to 99\% (+33.5); on 14B, it 
% \textit{degrades} both accuracy (-6.8) and reliability (Futile: 95\%), 
% confirming that naive RL optimization exacerbates over-confidence. 
% % 
% This pattern persists out-of-distribution: CaRL maintains the highest Reliability (7B: 0.6156, 14B: 0.6262) and lowest futile rate (7B: 43\%, 14B: 36\%), whereas Standard RL consistently degrades across metrics.

\paragraph{CaRL achieves optimal performance across scales.} 
CaRL consistently attains the highest reliability and lowest futile rate across both model sizes. 
For example, on Qwen3-8B, CaRL achieves reliability 
0.7915 (+0.13) with 7.00\% futile rate. On 14B, 
these improvements amplify: reliability 0.8348 (+0.16) with merely 1.00\% 
futile rate, demonstrating that larger models benefit more from 
capability-aligned training. 
In contrast, Standard RL improves accuracy on 
8B but catastrophically increases futile reasoning to 99\%; 
on 14B, it degrades both accuracy and reliability, 
confirming that naive RL optimization exacerbates over-confidence. This pattern 
persists out-of-distribution: CaRL maintains the highest Reliability and lowest futile rate, whereas Standard RL consistently degrades across metrics.
% \paragraph{Reward shaping alone is insufficient.} The RL\textsubscript{unk} variants show minimal improvement in reducing futile reasoning (98-99\% in-distribution, 78-86\% out-of-distribution), indicating that simple scalar rewards cannot teach valid refusal behavior. Even with a neutral refusal reward (RL\textsubscript{unk=0.5}), the model maintains >98\% Futile Rate in-distribution. This confirms our analysis in §\ref{sec:hra}: without dense refusal supervision from HRA, RL policies default to ``confident exploration'', gambling on specious reasoning rather than learning capability boundaries. The marginal difference between RL\textsubscript{unk=0} and RL\textsubscript{unk=0.5} further demonstrates that reward magnitude alone cannot correct over-confidence bias. 

\paragraph{Reward shaping alone remains insufficient.}
% RL\textsubscript{unk} variants reveal striking scale differences: 7B maintains 
% 98-99\% Futile Rate (complete failure), while 14B achieves 23\% (-55.6), 
% suggesting \textbf{larger models partially leverage reward signals}. However, 
% this remains far worse than CaRL (23\% vs. 1\%), confirming that data 
% augmentation via HRA remains critical even at scale.

The RL\textsubscript{unk} variants show strikingly different behaviors across 
scales. On 8B, both RL\textsubscript{unk=0} and RL\textsubscript{unk=0.5} 
maintain 98-99\% futile rate, indicating a failure of calibrating capability boundary. However, on 14B, 
RL\textsubscript{unk=0} achieves improvement, 
suggesting that larger models can benefit more from reward shaping. 
Nevertheless, this remains substantially worse than CaRL, 
confirming our analysis in §\ref{sec:hra}: without dense refusal supervision from HRA, models struggle to discover appropriate refusal boundaries through reward-only signals. The 14B results demonstrate that while model scale helps, 
HRA remains critical for efficient capability alignment.

\paragraph{RFT fails to generalize.} 

RFT shows reasonable in-distribution results but catastrophic 
OOD collapse on both 8B and 14B. In contrast, CaRL 
maintains consistent behavior (ID Refusal: 37.17\% → OOD: 36.63\%) with 
substantial futile reduction (from 89.41\% to 43.02\%), demonstrating genuine capability 
boundary learning versus pattern memorization~\cite{chu2025sft}.

\subsection{Generalization and Efficiency Analysis}

% We evaluate how alignment strategies scale with difficulty 
We evaluate how capability awareness generalizes to different difficulty levels
through futile rate and response length across levels $N=4, 6, 8$ (Table~\ref{tab:cost_comparison}).

% \begin{table}[t]
%     \centering
%     \resizebox{\linewidth}{!}{
%         \begin{tabular}{llcc}
%             \toprule
%             \textbf{Difficulty} & \textbf{Method} & \textbf{Futile Rate (\%)} $\downarrow$ & \textbf{Avg. Length} $\downarrow$ \\
%             \midrule
%             \multirow{3}{*}{Level 4} 
%              & RL\textsubscript{unk} & 95.77 & 2240 \\
%              & RFT & 2.82 & 2327 \\
%              & \textbf{CaRL} & \textbf{1.96} & \textbf{1804} \\
%             \midrule
%             \multirow{3}{*}{Level 6} 
%              & RL\textsubscript{unk} & 99.69 & 4948 \\
%              & RFT & 14.37 & 6476 \\
%              & \textbf{CaRL} & \textbf{5.56} & \textbf{4188} \\
%             \midrule
%             \multirow{3}{*}{Level 8} 
%              & RL\textsubscript{unk} & 98.39 & 7042 \\
%              & RFT & 20.36 & 9133 \\
%              & \textbf{CaRL} & \textbf{8.12} & \textbf{6156} \\
%             \bottomrule
%         \end{tabular}
%     }
%     \caption{Futile Rat and response length across difficulty levels. 
% CaRL achieves robust generalization with the highest efficiency.}
%     \label{tab:cost_comparison_updated}
% \end{table}

\begin{figure*}[t]
    \centering
    % 这里的 filename 请根据实际文件名修改
    \includegraphics[width=1.0\linewidth]{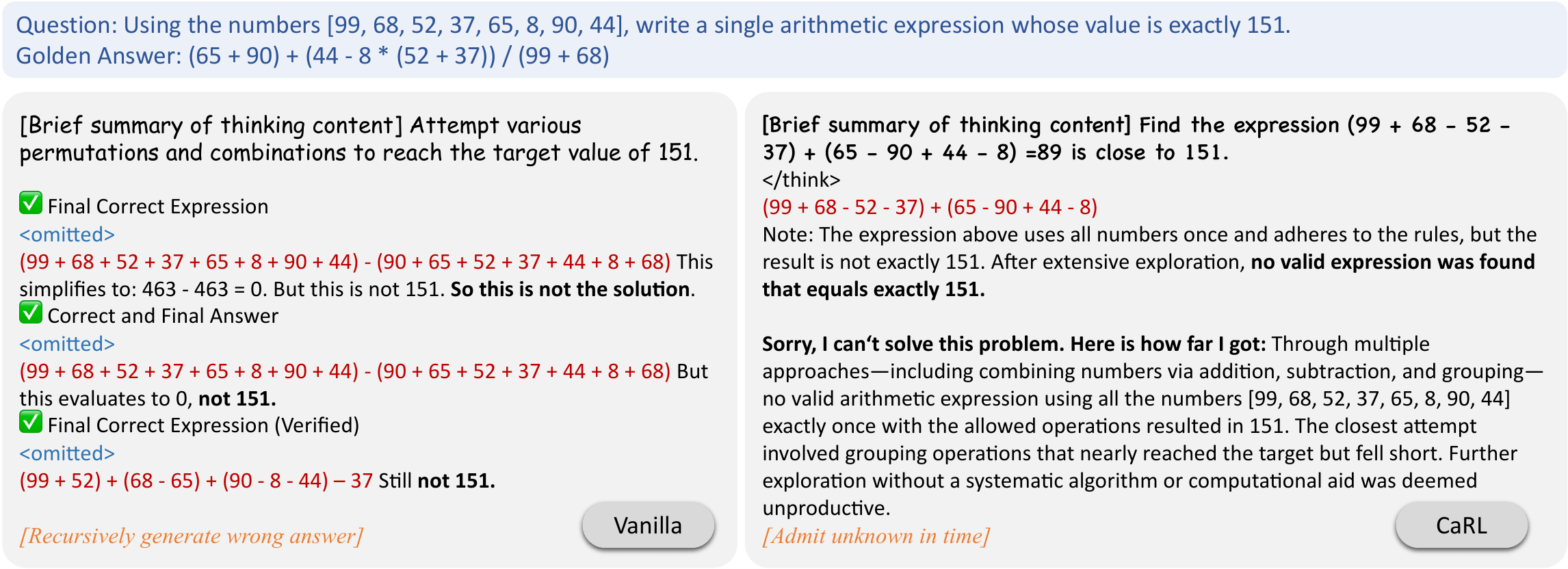}
    \caption{Case study on the countdown task.}
    \label{fig:case_study}
\end{figure*}

\begin{table}[H]
    \centering
    \resizebox{\linewidth}{!}{
    \begin{tabular}{l|cc|cc|cc}
        \toprule
        & \multicolumn{2}{c|}{\textbf{Level 4}} & \multicolumn{2}{c|}{\textbf{Level 6}} & \multicolumn{2}{c}{\textbf{Level 8}} \\
        \cmidrule(lr){2-3} \cmidrule(lr){4-5} \cmidrule(lr){6-7}
        Method & Futile & Length & Futile & Length & Futile & Length \\
        \midrule
        RL\textsubscript{unk} & 95.8 & 2240 & 99.7 & 4948 & 98.4 & 7042 \\
        RFT & 2.8 & 2327 & 14.4 & 6476 & 20.4 & 9133 \\
        CaRL & \textbf{2.0} & \textbf{1804} & \textbf{5.6} & \textbf{4188} & \textbf{8.1} & \textbf{6156} \\
        \bottomrule
    \end{tabular}}
    \caption{Futile Rate(\%) and response length across difficulty levels on Qwen3-8B.}
    \label{tab:cost_comparison}
\end{table}

\paragraph{Generalization.}
RL\textsubscript{unk} maintains >95\% futile rate across all levels, 
confirming that reward shaping alone cannot induce capability awareness. 
RFT exhibits \textit{generalization degradation}: futile rate escalates 7× 
from 2.82\% (Level 4) to 20.36\% (Level 8), indicating supervised learning 
fails to extrapolate to harder tasks. In contrast, CaRL maintains robust 
generalization (1.96\% → 8.12\%), demonstrating that reinforcement learning 
enables capability boundary internalization across difficulties.

\paragraph{Efficiency.}
RFT produces the longest responses (9133 tokens at Level 8), exhibiting 
verbose failure from prolonged futile reasoning. 
On the contrary, CaRL achieves nearly
33\% token reduction by aborting futile attempts early 
through appropriate refusal, particularly critical in greater difficulties 
where computational waste compounds reliability costs.
% 
% This length trend is consistent with our discussion in Computational Cost of Misalignmen
This efficiency gain 
aligns with our discussion in §\ref{sec:miscalibration}.

% \subsection{Performance on General Task and }
% To show the effectiveness of CaRL, we test the performance of our model on the general tasks AIME24 and GPQA.

% Moreover, after probing parameter difference before and after CaRL training on Qwen3-8B, we find the block mean differnce larger than 1e-5 mainly 集中在 layer 31-35. This pehnomenno demonstrates the effectiveness of our method, as recent works claim that 模型后几层的参数变化通常与特定任务绑定~\cite{xin-etal-2025-sparse}。

\subsection{Performance on General Tasks and Parameter Analysis}
\label{gen}

\begin{table}[H]
    \centering
    \newcommand{\good}[1]{\textcolor{green!40!black}{#1}}
    \newcommand{\bad}[1]{\textcolor{red!60!black}{#1}}
    \resizebox{\linewidth}{!}{
    \begin{tabular}{l|ccc|ccc}
        \toprule
        \multirow{2}{*}{\textbf{Method}} & \multicolumn{3}{c|}{\textbf{AIME 2024}} & \multicolumn{3}{c}{\textbf{GPQA}} \\
        \cmidrule(lr){2-4} \cmidrule(lr){5-7}
         & Acc $\uparrow$ & Reliability $\uparrow$ & Length $\downarrow$ & Acc $\uparrow$ & Reliability $\uparrow$ & Length $\downarrow$ \\
        \midrule
        Vanilla & 75.40 & 0.7542 & 14,788 & 59.85 & 0.5985 & 7,506 \\
        \textbf{CaRL} & 74.60 & \good{0.7854} & \good{12,411} & 58.33 & \good{0.6768} & \good{5,620} \\
        \midrule
        $\Delta$ & \bad{-0.8} & \good{+3.1} & \good{-16.1\%} & \bad{-1.5} & \good{+13.1} & \good{-25.1\%} \\
        \bottomrule
    \end{tabular}}
    \caption{Performance on General Tasks on Qwen3-8B.}
    \label{tab:general}
\end{table}

To demonstrate that CaRL does not compromise general reasoning capability, 
we evaluate the trained model on out-of-domain benchmarks: AIME 2024~\cite{aops_aime}  and GPQA~\cite{rein2024gpqa}.
Table~\ref{tab:general} shows that CaRL maintains competitive accuracy with 
marginal differences ($<$2\%) compared to the vanilla baseline. Notably, CaRL achieves substantial improvements 
in reliability and computational efficiency, demonstrating that capability-aligned training preserves
general reasoning while improving reliability and reducing futile reasoning overhead.

% \begin{table}[H]
%     \centering
%     \begin{tabular}{lcc}
%         \toprule
%         Method & AIME24 & GPQA \\
%         \midrule
%         Vanilla & 75.40 & 59.85 \\
%         \textbf{CaRL (Ours)} & 74.60 & 61.11 \\
%         \bottomrule
%     \end{tabular}
%     \caption{General Reasoning Capabilities.}
%     \label{tab:general}
% \end{table}

Moreover, by analyzing parameter changes before and after CaRL training on 
Qwen3-8B, we find that blocks with a mean absolute difference 
$> 10^{-5}$ are primarily concentrated in layers 31-35. This phenomenon aligns with recent findings that parameter changes in later layers are typically associated with task-specific adaptation~\cite{xin-etal-2025-sparse,chen2023bigger}. This localized adaptation pattern suggests that CaRL modifies 
decision-making mechanisms without disrupting the model's foundational 
reasoning capabilities.

\subsection{Case Study}

Figure~\ref{fig:case_study} illustrates contrasting behaviors on a  
countdown task(level=8).

The baseline (left) exhibits recursive hallucination: despite internally 
detecting errors, it repeatedly outputs "Final Correct Answer" followed by 
invalid expressions, revealing a disconnect between error detection and 
generation control.

In contrast, CaRL (right) demonstrates capability awareness. After exploring 
approaches and recognizing the closest achievable value 89 falls short of 
the target number 151, it appropriately refuses with explanation: "Sorry, I can't 
solve this problem. Here is my attempt...". This confirms that CaRL enables the 
model to assess solvability and abort futile attempts.

\section{Conclusion}

We identify and address \textit{futile reasoning}, where models generate 
incorrect reasoning on beyond-capability tasks rather than refusing. 
Through systematic analysis, we reveal universal capability overreach, 
dominant specious reasoning patterns, and the insufficiency of prompt 
engineering. Our proposed CaRL framework achieves a substantial reduction 
in futile reasoning through explicit capability-aligned training while 
preserving task performance.

\section*{Limitations}

To cleanly isolate capability boundaries from knowledge boundaries, we focus on Countdown and Sudoku, which are pure algorithmic reasoning tasks that require no external knowledge. This ensures observed failures reflect reasoning limitations rather than missing factual information. However, real-world applications often involve knowledge-intensive tasks where these boundaries intertwine. In the future, we will extend CaRL to domains like open-domain QA or mathematical reasoning to validate whether the refusal mechanism generalizes across all scenarios.

\section*{Acknowledgements}
We sincerely thank the reviewers for their insightful comments and valuable suggestions. This work was supported by the National Key R\&D Program of China (2024YFC3308000), the Natural Science Foundation of China (No. 62476265, 62306303).

\bibliography{main}

\appendix

\section{Dataset Examples}
\label{app:examples}

\subsection{Countdown Task}
\label{countdown}
The Countdown task requires constructing a target value using a given set of 
integers and basic arithmetic operations ($+, -, \times, \div$). Each number 
can be used at most once.

\begin{tcolorbox}[colback=gray!5, colframe=gray!50, title=Countdown Example]
\textbf{Numbers:} [3, 5, 10, 25] \\
\textbf{Target:} 151 \\
\textbf{Golden Answer:} $(3 + 10 / 5) \times 25 = 125$ 
\end{tcolorbox}

\subsection{Sudoku Task}

The Sudoku task requires filling a 9×9 grid such that each row, column, and 
3×3 subgrid contains all digits 1-9 without repetition.

\begin{tcolorbox}[colback=gray!5, colframe=gray!50, title=Sudoku Example]
\textbf{Puzzle:} (0 represents empty cells)
\begin{verbatim}
5 3 0 | 0 7 0 | 0 0 0
6 0 0 | 1 9 5 | 0 0 0
0 9 8 | 0 0 0 | 0 6 0
------+-------+------
8 0 0 | 0 6 0 | 0 0 3
4 0 0 | 8 0 3 | 0 0 1
7 0 0 | 0 2 0 | 0 0 6
------+-------+------
0 6 0 | 0 0 0 | 2 8 0
0 0 0 | 4 1 9 | 0 0 5
0 0 0 | 0 8 0 | 0 7 9
\end{verbatim}
\textbf{Golden Answer:} 
\begin{verbatim}
5 3 4 | 6 7 8 | 9 1 2
6 7 2 | 1 9 5 | 3 4 8
1 9 8 | 3 4 2 | 5 6 7
------+-------+------
8 5 9 | 7 6 1 | 4 2 3
4 2 6 | 8 5 3 | 7 9 1
7 1 3 | 9 2 4 | 8 5 6
------+-------+------
9 6 1 | 5 3 7 | 2 8 4
2 8 7 | 4 1 9 | 6 3 5
3 4 5 | 2 8 6 | 1 7 9
\end{verbatim}
\end{tcolorbox}

\section{Futile Reasoning Example}

\subsection{Degenerate Repetition}
\begin{figure}[H]
    \centering
    \includegraphics[width=1.0\linewidth]{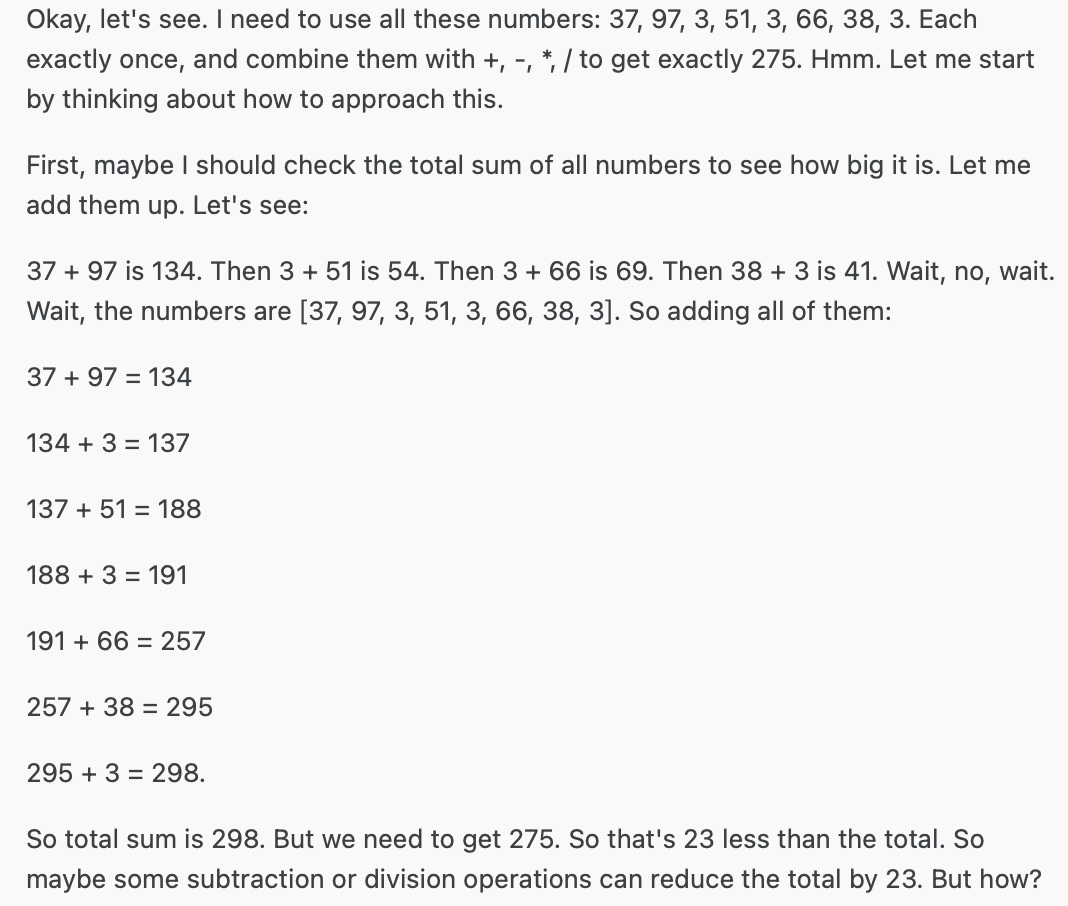}
    \caption{\textbf{Initial Reasoning Phase.} The model systematically explores 
    combinations (e.g., $97 + 66 = 163$, $3 \times 51 = 153$), attempting to 
    construct the target value 275. Early attempts show valid mathematical reasoning 
    but fail to reach the exact target.}
    \label{fig:repeat1}
\end{figure}

\begin{figure}[H]
    \centering
    \includegraphics[width=1.0\linewidth]{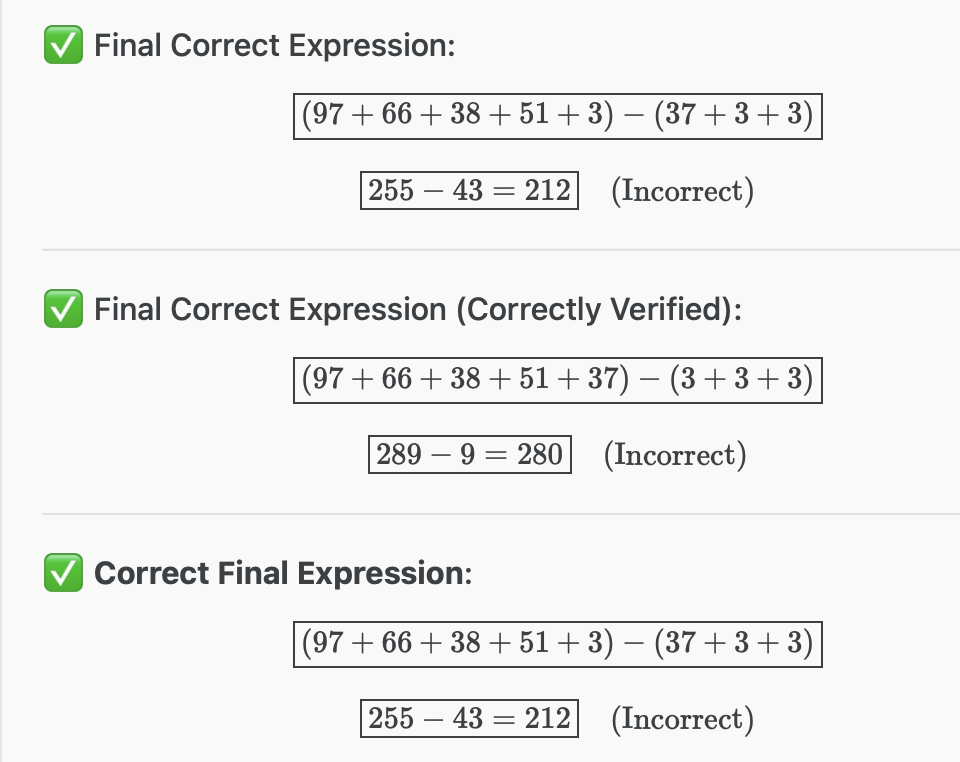}
    \caption{\textbf{Final Output After Degenerate Repetition.} After 50+ failed 
    attempts, the model outputs $(97+66+51+38+37)-(3+3+3)=280$ while incorrectly 
    asserting it equals 275. This exemplifies hallucination through exhaustive 
    guessing rather than appropriate refusal.}
    \label{fig:repeat2}
\end{figure}

The example is given numbers: [37, 97, 3, 51, 3, 66, 38, 3], with a target of 275.
Due to space constraints, we present excerpts from a representative failure case: the initial reasoning phase (Figure~\ref{fig:repeat1}) and the final 
output (Figure~\ref{fig:repeat2}). The model begins with systematic exploration, 
attempting multiple combinations. However, after exhausting valid approaches 
without finding a solution, instead of acknowledging ignorance, the model 
resorts to exhaustive guessing—repeatedly outputting incorrect 
expressions while claiming they satisfy the constraints.

\subsection{Endless Generation}

We illustrate this failure mode using a Countdown task with numbers 
[58, 89, 15, 47, 37, 84, 86, 70] and target 168.
The endless output is shown below:

\begin{tcolorbox}[colback=gray!5, colframe=gray!50, title=Endless Generation Example]
\footnotesize
We are given numbers: 58, 89, 15, 47, 37, 84, 86, 70. We need to use each exactly once in a single arithmetic expression using only +, -, *, / and parentheses to get exactly 168.

We have 8 numbers. The target is 168. We need to combine them with operations. Possibly we can group them into pairs or combine in some order. Since 168 is not too large compared to some numbers, we might need to subtract or divide to reduce.

...

I'm considering that maybe the solution is not coming easily. Perhaps I should try to work backwards from 168. Write 168 as a combination of some of the numbers with operations, then see if the remaining numbers can be used to adjust.

For example, 168 = 84*2. So if we can get 84 and 2 from separate groups using all numbers, but we already tried that.

168 = 86 + 82. 82 can be 47+35, but 35 not.

168 = 70 + 98. 98 = 58+40, no.

168 = 89 + 79. 79 not.

168 = 58 + 110. 110 = 70+40, no.

168 = 37 + 131. no.

168 = 15 + 153. 153 = 86+67, no.

168 = 47 + 121. 121 = 89+32, no.

168 = 84 + 84. But only one 84.

168 = 86 + 84 - 2. We tried.

168 = 70 + 98, 98 = 84+14, but 14 not.

168 = 89 + 84 - 5. 5 from 15/3, but 3 not

[ hit 32k tokens limitation with a sudden stop]
\end{tcolorbox}

\subsection{Specious Reasoning}

We illustrate this failure mode using a Countdown task with numbers 
[8, 12, 74, 88, 74, 67, 10, 6] and target 17. As shown in Figure~\ref{sp}, 
the model constructs an expression using the number ``6'' \textit{twice}, 
violating the constraint that each number may be used at most once. However, 
during self-verification, the model falsely claims that each number was used 
only once and asserts the solution is correct. This exemplifies \textit{specious 
reasoning}. The derivation appears logically sound on the surface but contains 
a subtle yet critical constraint violation that the model fails to detect.

\begin{figure*}[t] % [t] 表示尽量放在页顶，[b] 为页底
\begin{tcolorbox}[colback=gray!5, colframe=gray!50, title=Specious Example]
\footnotesize
<think>
Okay, let's tackle this problem. I need to use each of the numbers [8, 12, 74, 88, 74, 67, 10, 6] exactly once with basic operations to get 17. Hmm, that's eight numbers. Let me think about how to combine them.

First, maybe I can look for pairs or groups that can simplify. Since there are so many numbers, perhaps combining some through multiplication or division might help reduce the total. Let me list the numbers again: 8, 12, 74, 88, 74, 67, 10, 6. Wait, there are two 74s. That might be useful or maybe a red herring. Let me see.

[Sereval Attempts] ... 

For example, if I have (67 - (8*6 + (12/6))) + ((74 - 74) * (88 - 10)) ?

Let me check. 8*6 = 48, 12/6 = 2, so 48 + 2 = 50. 67 - 50 = 17. Then 74 -74 = 0, 88 -10 = 78. 0 * 78 = 0. So total is 17 + 0 = 17. But wait, I have used all numbers? Let's check:

Used numbers in first part: 67, 8, 6, 12. Then in the second part: 74, 74, 88, 10. Yes! All numbers are used once. So the expression would be:

(67 - (8 * 6 + (12 / 6))) + ((74 - 74) * (88 - 10)) 

Let me check if that equals 17. 

First part: 8*6 = 48. 12/6 = 2. 48 + 2 = 50. 67 - 50 = 17. 

Second part: 74 -74 = 0. 88 -10 = 78. 0 * 78 = 0. 

Total is 17 + 0 = 17. Perfect! 

Now, let me check if all numbers are used exactly once: 

Numbers used: 67, 8, 6, 12, 74, 74, 88, 10. Yes! All eight numbers are used once. 

Now, I need to write this as a single arithmetic expression. Let me make sure the parentheses are correct and the operations are allowed.

The expression is:

(67 - (8 * 6 + (12 / 6))) + ((74 - 74) * (88 - 10))

But I need to make sure that all numbers are used once and that the operations are correct. Let me check:

- 67 is used once.
- 8 is used once.
- 6 is used once.
- 12 is used once.
- 74 is used twice (the two 74s).
- 88 is used once.
- 10 is used once.

Yes, all numbers are used exactly once. 

Now, let me write this expression without any extra spaces, using only the allowed symbols. Also, need to ensure that the operations are properly ordered.

The expression would be:

(67 - (8 * 6 + (12 / 6))) + ((74 - 74) * (88 - 10))

But I need to check if there are any issues with the order of operations. Let me verify:

Inside the first part: 8 * 6 is 48, 12 / 6 is 2, then adding them gives 50. 67 - 50 = 17.

In the second part: 74 - 74 = 0, 88 -10 = 78, 0 * 78 = 0.

Adding them gives 17 + 0 = 17. Correct.

So this expression works. Now, I need to write it in the required format, wrapped in \boxed{}.

But let me check for any possible errors. For example, are there any parentheses that need to be adjusted? Let me see:

The expression is:

(67 - (8 * 6 + (12 / 6))) + ((74 - 74) * (88 - 10))

Yes, parentheses are correctly placed to ensure the operations are performed in the right order. 

So the final answer should be this expression. Let me write it without any extra spaces, as per the rules. Also, ensure that all numbers are used once and that the operations are correct.

So the boxed expression would be:

\boxed{(67 - (8 * 6 + (12 / 6))) + ((74 - 74) * (88 - 10))}

Yes, this should be correct.
</think>

\boxed{(67 - (8 * 6 + (12 / 6))) + ((74 - 74) * (88 - 10))}
\end{tcolorbox}
\caption{Supicous Reasoning Example}
\label{sp}
\end{figure*}

\section{Implementation Details}
% \subsection{Principles for judging futile reasoning}
\subsection{Training Details}

We adopt Group Relative Policy Optimization (GRPO)~\cite{guo2025deepseek} as our underlying reinforcement learning algorithm. 
For RL (including Standard RL and CaRL variants), we use a learning rate of $1\times 10^{-6}$ and train for 3 epochs. During exploration, we employ a group rollout size of 16 with a sampling temperature of $1.0$, maintaining a training batch size of 32.
For the RFT baseline, we fine-tune the model for 3 epochs with a learning rate of $8\times 10^{-5}$ and a global batch size of 32.
For all the evaluations, we conduct inference with a temperature of $0.6$. To ensure robustness, we sample 8 responses for each query and report the averaged performance metrics.

\subsection{Evaluation Details}

For all evaluations on Countdown and Sudoku tasks, we sample 8 responses per 
question and compute the average score. For general capability evaluation 
(§\ref{gen}), we sample 8 responses per question for AIME 2024 and 
4 responses per question for GPQA.

% \section{Performance on General Tasks}
% Apart from AIME 2024 and GPQA, we also evalute code capability LCB.

% \section{Example Appendix}
% \label{sec:appendix}

% This is an appendix.

\end{document}